\documentclass[runningheads]{llncs}

\usepackage{eccv}

\usepackage{eccvabbrv}

\usepackage{graphicx}
\usepackage{booktabs}

\usepackage[accsupp]{axessibility}  

\usepackage[pagebackref,breaklinks,colorlinks,citecolor=eccvblue]{hyperref}

\usepackage{orcidlink}

\begin{document}

\title{DreamScene360: Unconstrained Text-to-3D Scene Generation with Panoramic Gaussian Splatting} 

\titlerunning{DreamScene360}

\author{Shijie Zhou\inst{1}$^\star$\orcidlink{0000-0002-9018-7539} \and
Zhiwen Fan\inst{2}\thanks{Equal contribution.}\orcidlink{0000-0002-8302-7465
} \and
Dejia Xu\inst{2}$^\star$\orcidlink{0000-0001-8474-3095} \and
Haoran Chang\inst{1}\orcidlink{0009-0005-6333-239X} \and \\
Pradyumna Chari\inst{1}\orcidlink{0000-0002-9610-0350} \and
Tejas Bharadwaj\inst{1}\orcidlink{0009-0007-2370-8510} \and
Suya You\inst{3}\orcidlink{0000-0002-6387-7024} \and \\
Zhangyang Wang\inst{2}\orcidlink{0000-0002-2050-5693} \and
Achuta Kadambi\inst{1}\orcidlink{0000-0002-2444-2503}
}

\authorrunning{S.~Zhou et al.}


\institute{University of California, Los Angeles \and
University of Texas at Austin \and
DEVCOM Army Research Laboratory \\
\url{http://dreamscene360.github.io/}
}

\maketitle

\begin{figure}[h]
    \centering
    \includegraphics[width=0.864\textwidth]{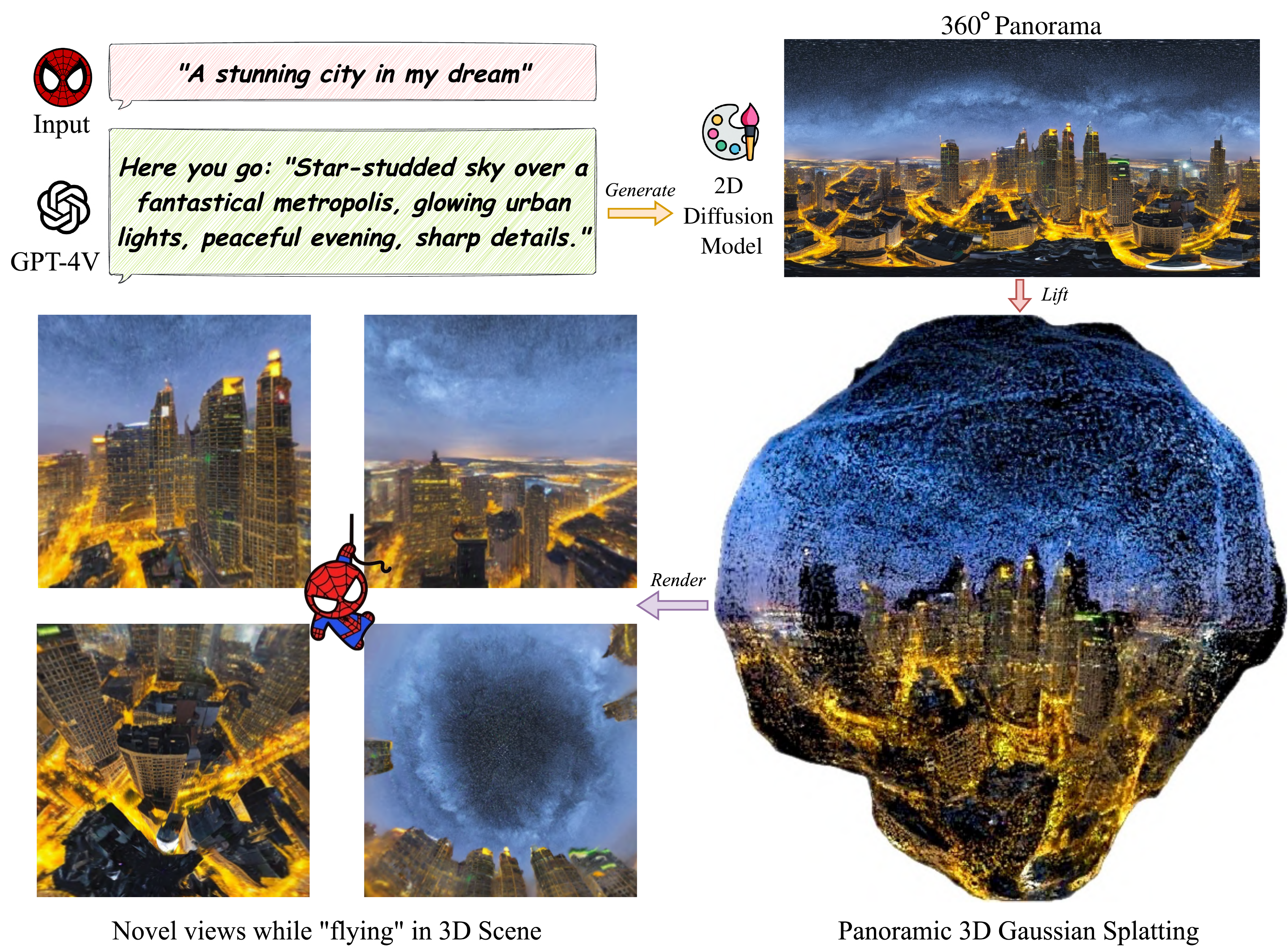}    
\caption{\textbf{DreamScene360.} We introduce a 3D scene generation pipeline that creates immersive scenes with full 360$^{\circ}$ coverage from text prompts of any level of specificity.}
\label{fig:teaser}
\end{figure}

\begin{abstract}

The increasing demand for virtual reality applications has highlighted the significance of crafting immersive 3D assets. We present a text-to-3D 360$^{\circ}$ scene generation pipeline that facilitates the creation of comprehensive 360$^{\circ}$ scenes for in-the-wild environments in a matter of minutes. Our approach utilizes the generative power of a 2D diffusion model and prompt self-refinement to create a high-quality and globally coherent panoramic image. This image acts as a preliminary ``flat'' (2D) scene representation. Subsequently, it is lifted into 3D Gaussians, employing splatting techniques to enable real-time exploration. To produce consistent 3D geometry, our pipeline constructs a spatially coherent structure by aligning the 2D monocular depth into a globally optimized point cloud. This point cloud serves as the initial state for the centroids of 3D Gaussians. In order to address invisible issues inherent in single-view inputs, we impose semantic and geometric constraints on both synthesized and input camera views as regularizations. 
These guide the optimization of Gaussians, aiding in the reconstruction of unseen regions. In summary, our method offers a globally consistent 3D scene within a 360$^{\circ}$ perspective, providing an enhanced immersive experience over existing techniques.
Project website at: \url{http://dreamscene360.github.io/}.

\end{abstract}

\section{Introduction}
\label{sec:intro}
The vast potential applications of text-to-3D to VR/MR platforms, industrial design, and gaming sectors have significantly propelled research efforts aimed at developing a reliable method for immersive scene content creation at scale.
Recent developments in the 2D domain have seen the successful generation or editing of high-quality and adaptable images/videos using large-scale pre-trained diffusion models~\cite{rombach2022high,ramesh2022hierarchical} on large-scale datasets, allowing users to generate customized content on demand.

Moving beyond 2D, the generation of 3D content, particularly 3D scenes, is constrained by the limited availability of annotated 3D image-text data pairs. Consequently, efforts in 3D content creation often rely on leveraging large-scale 2D models. 
This line of approach facilitates the creation of 3D scenes through a time-consuming distillation process. An example of this is DreamFusion~\cite{poole2022dreamfusion}, which seeks to distill the object-wise 2D priors from diffusion models into a 3D neural radiance field (NeRF)~\cite{mildenhall2020nerf}.
However, these approaches often suffer from low rendering quality, primarily due to the multi-view inconsistency of 2D models, and struggle to extend to scene-scale 3D structure with fine details texture creation, particularly for outdoor scenes~\cite{poole2022dreamfusion} with outward-facing viewpoints and unbounded scene scale.
Another avenue of 3D generation draws insights from explicit representations, such as point clouds and meshes, as demonstrated in LucidDreamer~\cite{chung2023luciddreamer} and Text2Room~\cite{hollein2023text2room}. These methods attempt to bridge the gap between 2D and 3D generation by initializing with an explicit 3D representation, and then progressively expanding the learned 3D representation to  encompass a broader field-of-view. 
However, the progressive optimization frameworks leveraged by these methods struggle to inpaint substantial missing areas, especially when targeting 360$^\circ$ scenes under unconstrained conditions, resulting in notably distorted and disjointed structures.
Moreover, the issue of prompt engineering in text-to-image generation~\cite{rombach2022high,imagen}, becomes more pronounced in text-to-3D generation frameworks~\cite{chung2023luciddreamer,poole2022dreamfusion,armandpour2023re} that rely on either time-consuming score distillation or complex, multi-step progressive inpainting during the scene generation process, leading to a considerable trial-and-error effort to achieve the desired 3D scene.


To address the above challenges in creating a holistic 360$^\circ$ text-to-3D scene generation pipeline, we introduce \textbf{DreamScene360}. 
Our method initially leverages the generative capabilities of text-to-panorama diffusion models~\cite{wang2024customizing} to produce omnidirectional 360$^\circ$ panoramas providing a comprehensive representation of the scene. A self-refining mechanism is used to enhance the image to alleviate prompt engineering, where GPT-4V is integrated to improve the visual quality and the text-image alignment through iterative quality assessment and prompt revision. 
While the generated panorama images overcome the view consistency issue across different viewpoints, they still lack depth information and any layout priors in unconstrained settings, and contains partial observations due to their single-view nature. 
To address this, our approach involves initializing scale-consistent scene geometry by employing a pretrained monocular depth estimator alongside an optimizable geometric field, facilitating deformable alignment for each perspective-projected pixel.
The gaps, stemming from single-view observations, can be filled by deforming the Gaussians to the unseen regions by creating a set of pseudo-views with a synthesized multi-view effect and the distillation of pseudo geometric and semantic constraints from 2D models (DPT~\cite{ranftl2021vision} and DINOv2~\cite{oquab2023dinov2}) to guide the deformation process to alleviate artifacts.

Collectively, our framework, DreamScene360, enables the creation of immersive and realistic 3D environments from a simple user command, offering a novel solution to the pressing demand for high-quality 3D scenes (see the workflow in Fig.~\ref{fig:teaser}). Our work also paves the way for more accessible and user-friendly 3D scene generation by reducing the reliance on extensive manual effort.

\section{Related Works}
\paragraph{\textbf{2D Assets Generation.}}
The generation of 2D assets allows for incredible creative liberty, and the use of large-scale learning based priors for content generation. Generative Adversarial Networks~\cite{goodfellow2020generative} were the original state of the art for image generation. Variants such as StyleGAN~\cite{karras2019style} showed the ability for fine-grained control of attributes such as expression through manipulation of the latent representations as well. More recently, denoising diffusion models~\cite{dhariwal2021diffusion,ho2020denoising,song2020denoising} have been the new state of the art for generative models. Text-guided image diffusion models~\cite{saharia2022photorealistic} have shown the capability of generating high-quality images that are faithful to the conditioning text prompts. Subsequent work has made the generation process more efficient by performing denoising in the latent space~\cite{rombach2022high}, and by speeding up the denoising process~\cite{salimans2022progressive,meng2023distillation,geng2024one}. Techniques such as classifier free guidance~\cite{ho2022classifier} have significantly improved faithfulness to text prompts. Additional control of generation has also been shown to be possible, through auxiliary inputs such as layout~\cite{zheng2023layoutdiffusion}, pose~\cite{zhang2023adding} and depth maps~\cite{bhat2023loosecontrol}. More recently, text-to-image diffusion models have been finetuned to generate structured images, such as panoramas~\cite{wang2024customizing}. In this work, we utilize text-to-panorama generation as structured guidance for text to 360$^\circ$ scene generation.

\paragraph{\textbf{Text-to-3D Scene Generation.}}
In recent times, text-to-3D scene generation has been reinvigorated with the advent of 2D diffusion models and guidance techniques. Text-to-image diffusion models, along with techniques such as classifier free guidance, have been found to provide strong priors to guide 3D generation methods~\cite{poole2022dreamfusion,wang2023prolificdreamer,wang2023score,song2023roomdreamer,xu2022neurallift}. These methods are generally more frequently used than direct text-to-3D diffusion models~\cite{jun2023shap,karnewar2023holodiffusion}. More recent works~\cite{lin2023componerf,gao2023graphdreamer,vilesov2023cg3d} successfully generate multi-object compositional 3D scenes. On the other hand, a second class of methods use auxiliary inputs such as layouts~\cite{po2023compositional}. The least constrained text-to-3D methods do not rely on any auxiliary input with the only input being the text prompt describing the 3D scene~\cite{hollein2023text2room,zhang2024text2nerf,ouyang2023text2immersion,fang2023ctrl,mao2023showroom3d,yu2023wonderjourney,chung2023luciddreamer}. Our work requires a text prompt input; however, unlike prior work, we propose using panoramic images as an intermediate input for globally consistent scenes.

\paragraph{\textbf{Efficient 3D Scene Representation.}}
3D scene representations include a wide variety of techniques, including point clouds~\cite{berger2014state}, volumetric representations~\cite{lombardi2019neural,nguyen2019hologan}, and meshes~\cite{zhang2001efficient}. More recently, however, learning-based scene representations have gained prominence. Implicit representations such as neural radiance fields~\cite{mildenhall2021nerf} have shown high quality rendering and novel view synthesis capability. Additional learning-based representations, that span both explicit~\cite{fridovich2022plenoxels} and mixed~\cite{muller2022instant} have been proposed to speed up learning. Most recently, 3D Gaussian splatting~\cite{kerbl20233d} has enabled fast learning and rendering of high quality radiance fields using a Gaussian kernel-based explicit 3D representation. Since then, several works have emerged enabling sparse view~\cite{zhu2023fsgs,xiong2023sparsegs} and compressed~\cite{niedermayr2023compressed,fan2023lightgaussian,navaneet2023compact3d,lee2023compact,morgenstern2023compact} 3D Gaussian representations, as well as representation of multidimensional feature fields~\cite{zhou2024feature}. A variation involves learning a Gaussian-based radiance field representation using panoramic images as input~\cite{bai2024360}. In this work, we propose a method for text to 360$^\circ$ 3D scene generation, by using panorama images as an intermediate representation.

\section{Methods}
\begin{figure}[t]
  \centering
  \includegraphics[width=1.0\linewidth]{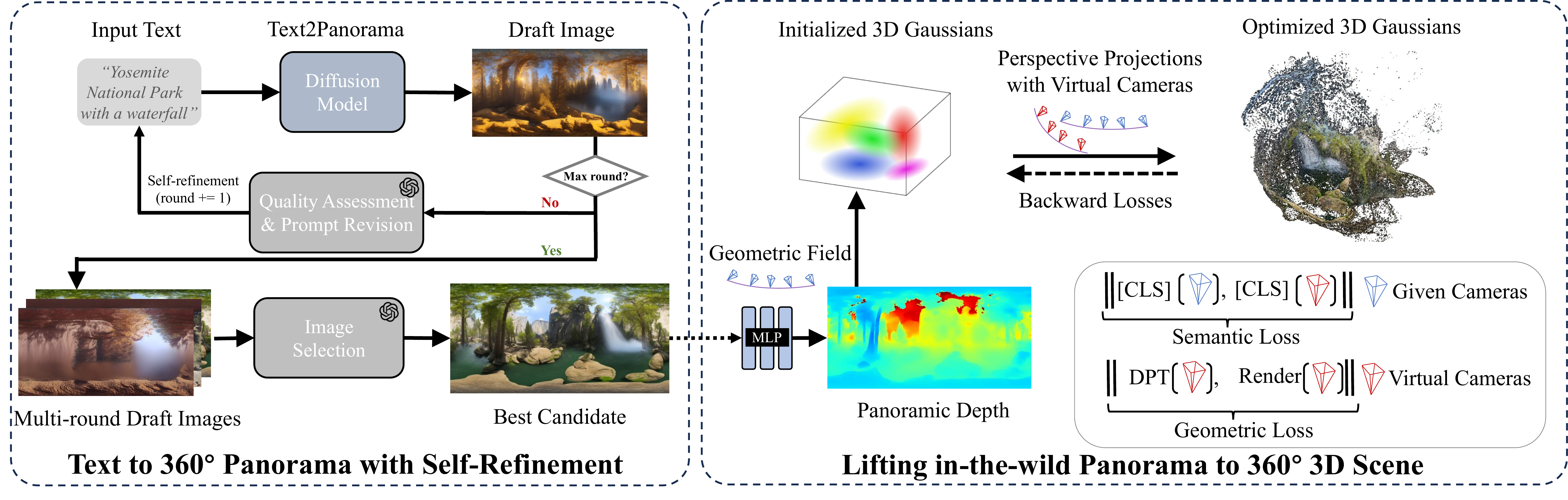}
  \caption{\textbf{Overall Architecture.} Beginning with a concise text prompt, we employ a diffusion model to generate a 360$^\circ$ panoramic image. A self-refinement process is employed to produce the optimal 2D candidate panorama. Subsequently, a 3D geometric field is utilized to initialize the Panoramic 3D Gaussians. Throughout this process, both semantic and geometric correspondences are employed as guiding principles for the optimization of the Gaussians, aiming to address and fill the gaps resulting from the single-view input.}
  \label{fig:arc}
\end{figure}

In this section, we detail the proposed DreamScene360 architecture (Fig.\ref{fig:arc}). DreamScene360 initially generates a 360$^\circ$ panorama utilizing a self-refinement process, which ensures robust generation and aligns the image with the text semantically (Sec.~\ref{sec:text2img}). The transformation from a flat 2D image to a 3D model begins with the initialization from a panoramic geometric field (Sec.~\ref{sec:img23D}), followed by employing semantic alignment and geometric correspondences as regularizations for the deformation of the 3D Gaussians (Sec.~\ref{sec:3dgs_refine}).

\subsection{Text to 360$^\circ$ Panoramas with Self-Refinement}\label{sec:text2img}
Panoramic images provide an overview of the entire scene in one image, which is essential for generating 360$^\circ$ 3D scenes with global consistency. 

\paragraph{\textbf{360$^\circ$ Panoramic Image Generation.}}
A crucial requirement for the generated panorama is ensuring continuity between the leftmost and rightmost sides of the image.
We utilize the diffusion process from MultiDiffuser~\cite{bar2023multidiffusion} to generate a panoramic image ${I_0}$ of resolution $H \times 2H$ based on a pre-trained diffusion model $\Phi$.
Starting from a noisy image ${I}_T$, we iteratively denoise the image by solving an optimization problem for each of several patches of the image, selected via a sliding window mechanism. For each patch $P_i(I_t)$, we ensure the distance against their denoised version $\Phi(P_i(I_t))$ is minimized. Though each patch may pull the denoising process in a different direction, the fused result is the weighted average of the update of each sample, whose closed form is written as follows,
\begin{equation}\label{eq:closed-form}
\Phi(I_{t-1}) = \sum_{i=1}^{n}\frac{P^{-1}_i(W_i)}{\sum_{j=1}^{n}P_{j}^{-1}(W_j)} \otimes P_i^{-1}(\Phi(P_i(I_t))),
\end{equation}
where $W_i$ refers to per pixel weight, set to 1 in our experiments.



We use StitchDiffusion~\cite{wang2024customizing} as the pretrained 2D diffusion model, where a stitch method is employed in the generation process for synthesizing seamless 360$^\circ$ panoramic images. Trained on a curated paired image-text dataset containing 360$^\circ$ panoramas, a customized LoRA~\cite{hu2021lora} module is incorporated into the MultiDiffusion~\cite{bar2023multidiffusion} process.
At each denoising timestamp, we not only diffuse at the original resolution $H \times (2H + 2W)$ via MultiDiffuser $\Phi$, but also stitch the leftmost and rightmost image regions ($H\times W$ each) and diffuse the concatenated patch as well to ensure consistency at the border regions. 
Finally, we consider the center cropped region of $H \times 2H$ as the final 360$^\circ$ panoramic image. It is worth noting that Dreamscene360 is versatile in practice and can also adapt to other text-to-panorama diffusion models.

\paragraph{\textbf{Multi-Round Self-Refinement.}}
Unlike previous works that generate 3D scenes through a time-consuming score distillation process~\cite{wang2023prolificdreamer} or progressive inpainting~\cite{chung2023luciddreamer}, our work uses panorama to achieve user-friendly ``one-click'' 3D scene generation. We integrate GPT-4V to facilitate iterative self-refinement during the generation process, a feature that was challenging to implement in previous baselines due to the absence of global 2D image representations.
Here, we draw inspiration from perspective image generation, Idea2Img~\cite{yang2023idea2img}, and implement a self-improvement framework aiming for better text-image aligned panoramic image generation. 
Starting from a user-provided rough prompt, we leverage GPT-4V to provide feedback and prompt revision suggestions according to the results generated by the previous step. 
During each round, GPT-4V is judges the generated image quality in terms of object counts, attributes, entities, relationships, sizes, appearance, and overall similarity with the original user-specified prompt.
A score from 0-10 is assigned to each draft image and the one image with highest score is provided to GPT-4V for additional improvement.
GPT-4V will then produce an improved text prompt based on the issues observed in the current generation results, and the new prompt will be used for another round of panorama generation.
After a number of rounds, we collect the image with the highest visual quality score judged by GPT-4V in the whole generation process as our final panorama image. 

Our 3D scene generation framework, therefore, enjoys a user-friendly self-improvement process, without the need for troublesome prompt engineering for users as in previous methods~\cite{poole2022dreamfusion,wang2023prolificdreamer,chung2023luciddreamer}, but is able to obtain a high-quality, visually pleasing, text-aligned, and 360$^\circ$ consistent panoramic images that can be later converted into an immersive 3D scene via Panoramic Gaussian Splatting, which is detailed in the next sections.



\subsection{Lifting in-the-wild Panorama to 360 Scene}\label{sec:img23D}
Transforming a single image, specifically an in-the-wild 360$^\circ$ panoramic image, into a 3D model poses significant challenges due to inadequate observational data to regularize the optimization process, such as those required in 3D Gaussian Splatting (see Fig.~\ref{fig:init}).
Rather than beginning with a sparse point cloud (3DGS), we initialize with a dense point cloud utilizing pixel-wise depth information from the panoramic image of resolution  H $\times$ W, which are then refined towards more precise spatial configurations, ensuring the creation of globally consistent 3D representations, robust to viewpoint changes.
\paragraph{\textbf{Monocular Geometric Initialization.}}
Given a single panoramic image  $\mathbf{P}$, we project it onto N perspective tangent images with overlaps $\{ (\boldsymbol{I}_i  \in \mathbb{R}^{H \times W \times 3}, \boldsymbol{P}_i \in \mathbb{R}^{3 \times 4}) \}_{i=1}^{N}$, following the literature's suggestion that 20 tangent images adequately cover the sphere's surface as projected by an icosahedron~\cite{rey2022360monodepth}. Unlike indoor panoramas, which benefit from structural layout priors in optimization, we employ a monocular depth estimator, DPT~\cite{ranftl2021vision}, to generate a monocular depth map $\boldsymbol{D}^{\text{Mono}}_i$ providing a robust geometric relationship. Nevertheless, these estimators still possess affine ambiguity, lacking a known scale and shift relative to metric depth. Addressing this efficiently is crucial for precise geometric initialization.
\paragraph{\textbf{Global Structure Alignment.}}
Previous studies in deformable depth alignment~\cite{hedman2018instant,wang2023perf,rey2022360monodepth} and pose-free novel view synthesis~\cite{bian2023nope} have explored aligning scales across multiple view depth maps derived from monocular depth estimations. In this context, we utilize a learnable global geometric field (MLPs), inspired by~\cite{wang2023perf, rey2022360monodepth}, supplemented by per-view scale and per-pixel shift parameters: $\{ (\alpha_i  \in \mathbb{R}, \boldsymbol{\beta_i} \in \mathbb{R}^{H\times W} \}_{i=1}^{N}$.
We define the view direction ($\boldsymbol{v}$) for all pixels on the perspective images deterministically. The parameters of MLPs $\Theta$ are initialized with an input dimension of three and an output dimension of one. The parameters $\{ (\alpha_i  \in \mathbb{R} \}_{i=1}^{N}$ are initialized to ones, and $\{ (\boldsymbol{\beta_i}  \in \mathbb{R}^{H\times W} \}_{i=1}^{N}$ to zeros. With these optimizable parameters, we define our optimization goal as follows:


\begin{align}
& \min_{\alpha, \beta, \Theta} \bigg\{ ||  \alpha \cdot \boldsymbol{D}^{\text{Mono}} +  \boldsymbol{\beta}  - \operatorname{MLPs}(\boldsymbol{v}; \Theta) ||_2^2 + \lambda_{\text{TV}} \mathcal{L}_{\text{TV}}(\boldsymbol{\beta}) + \lambda_{\alpha} ||\gamma (\alpha) - 1||^2 \bigg\} \\  \nonumber
& \text{where} \quad \mathcal{L}_{\text{TV}}(\boldsymbol{\boldsymbol{\beta}}) = \sum_{i,j} \left( (\beta_{i,j+1} - \beta_{i,j})^2 + (\beta_{i+1,j} - \beta_{i,j})^2 \right)
\end{align}

Here,  $\lambda_{\text{TV}}$ and $\lambda_{\alpha}$ are regularization coefficients to balance the loss weight, $\gamma(\cdot)$ is the softplus function. We set FoV as 80$^\circ$ during the optimization and use the predicted depth from the MLPs for the subsequent Gaussian optimization, as it can provide the depth of any view direction.


\subsection{Optimizing Monocular Panoramic 3D Gaussians}\label{sec:3dgs_refine}
While 3D Gaussians initialized with geometric priors from monocular depth maps provide a foundational structure, they are inherently limited by the lack of parallax inherent to single-view panoramas. This absence of parallax — critical for depth perception through binocular disparity — along with the lack of multiple observational cues typically provided by a baseline, poses substantial challenges in accurately determining spatial relationships and depth consistency. Thus, it is imperative to employ more sophisticated and efficient generative priors that function well in unconstrained settings and enable the scaling up and extend to any text prompts.
\paragraph{\textbf{3D Gaussian Splatting.}}
3D Gaussian Splatting (3DGS) utilizes the multi-view calibrated images using Structure-from-Motion~\cite{schonberger2016structure}, and optimizes a set of Gaussians with center $\boldsymbol{x} \in \mathbb{R}^3 $, an opacity value $\alpha \in \mathbb{R}$, spherical harmonics (SH) coefficients $\boldsymbol{c} \in \mathbb{R}^C$, a scaling vector $\boldsymbol{s} \in \mathbb{R}^3 $ and a rotation vector $\boldsymbol{q} \in \mathbb{R}^4 $ represented by a quaternion.
Upon projecting the 3D Gaussians into a 2D space, the color $C$ of a pixel is computed by volumetric rendering, which is performed using front-to-back depth order~\cite{kopanas2021point}: 
\begin{equation}
C = \sum_{i \in \mathcal{N}} c_i \alpha_iT_i, 
\label{eq:frender}
\end{equation}
where $T_i = \prod_{j=1}^{i-1} (1 - \alpha_j)$,  $\mathcal{N}$ is the set of sorted Gaussians overlapping with the given pixel, $T_i$ is the transmittance, defined as the product of opacity values of previous Gaussians overlapping the same pixel. 

\begin{figure}[htbp]
    \centering
    \includegraphics[width=\textwidth]{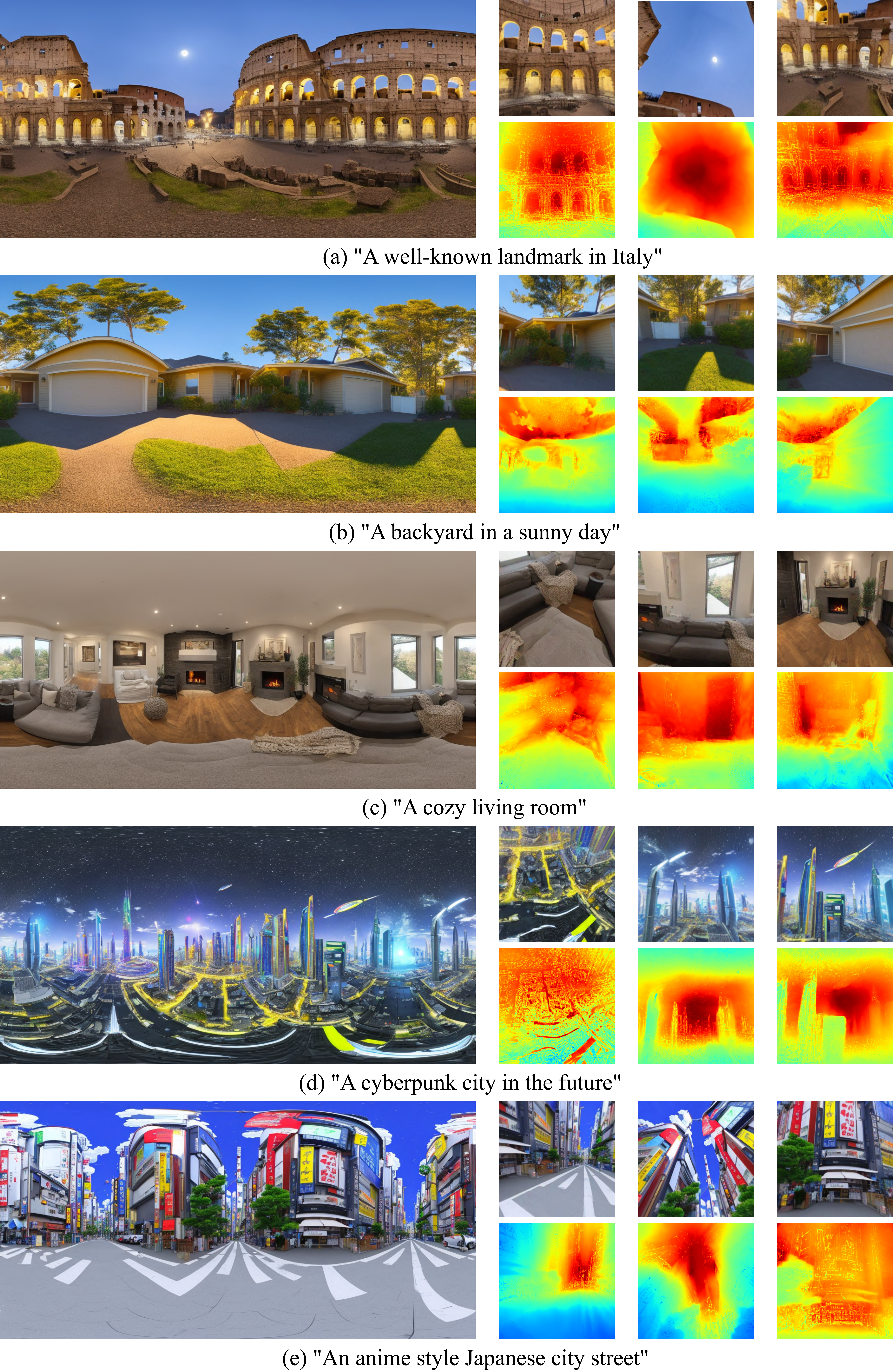}    
\caption{\textbf{Diverse Generation.} We demonstrate that our generated 3D scenes are diverse in style, consistent in geometry, and highly matched with the simple text inputs.}
\label{fig:diverse}
\end{figure}

\paragraph{\textbf{Synthesize Parallax with Virtual Cameras.}}
We emulate parallax by synthesizing virtual cameras that are unseen in training but are close to the input panoramic viewpoint. We methodically create these cameras to simulate larger movements. This procedure is quantitatively described by incrementally introducing perturbation to the panoramic viewpoint coordinates $(x, y, z)$ formalized as:
\begin{equation}\label{eq:virtual_cam}
    (x', y', z') = (x, y, z) + \delta(d_x, d_y, d_z)
\end{equation}
where $(x', y', z')$ denotes the new virtual camera positions, and $\delta(d_x, d_y, d_z)$ denotes the progressively increasing perturbations in each coordinate direction, under uniform distribution over $[-0.05, +0.05] \times \gamma$, where $\gamma \in \{1, 2, 4\}$ stands for 3-stage progressive perturbations, emulating the camera's movement from the original point.

\paragraph{\textbf{Distilling Semantic Similarities.}}
Previous research has underscored the significance of capturing the appearance of objects/scenes to establish robust feature correspondences across different views~\cite{henaff2022object,li2018deep}, which is crucial for tasks like co-segmentation and 2D pre-training. Inspired by these studies, we aim to establish a connection in visual feature correspondence between training views and synthetically generated virtual views. This is achieved by enforcing feature-level similarity, guiding the 3D Gaussians to fill the geometric gaps in the invisible regions effectively.
To elaborate, we generate a perspective image $\boldsymbol{I}_i$ from the panoramic 3D Gaussians at a specific training camera viewpoint $\boldsymbol{P}_i$ and create another synthesized image $\boldsymbol{I}'_i$ using Eq~\ref{eq:virtual_cam} that is proximate to $\boldsymbol{P}_i$. We employ the [CLS] token from the pre-trained DINOv2~\cite{oquab2023dinov2} model to encapsulate compact semantic features through the equation:
\begin{equation}
    \mathcal{L}_{\textit{sem}} = 1 - \operatorname{Cos}(\text{[CLS]}(I_i), \text{[CLS]}(I'_i)),
\end{equation}
where $\operatorname{Cos}$ denotes the cosine similarity. This approach is inspired by~\cite{tumanyan2022splicing,xu2022sinnerf}, assuming that the [CLS] token from a self-supervised, pre-trained Vision Transformer (ViT) can capture the high-level semantic attributes of an image. Such a mechanism is instrumental in identifying and leveraging similarities between two adjacent rendered images, facilitating our end-to-end 3DGS optimization process.

\paragraph{\textbf{Regularizing Geometric Correspondences.}}
While leveraging appearance cues is valuable, relying solely on them can lead to spatial inconsistencies. This is because models like DINOv2 might prioritize capturing semantic features over maintaining geometric coherence, potentially resulting in artifacts such as floaters. To mitigate this, we introduce a geometric regularization strategy designed to penalize discontinuities between pixels that exhibit inaccurate depth relationships. We employ a monocular depth estimator DPT~\cite{ranftl2021vision} for this purpose~\cite{deng2023nerdi}. Although it may not provide globally accurate depth, it allows us to ascertain the relative spatial relationships between pixels:
\begin{equation}
    \mathcal{L}_\textit{geo}(I_i, D_i) = 1 - \frac{\text{Cov}(D_i, \text{DPT}(I_i))}{\sqrt{\text{Var}(D_i)\text{Var}(\text{DPT}(I_i))}},
\end{equation}
Here, $I_i$ represents the rendered image at the $i$-th camera, and $D_i$ signifies the rendered depth. Additionally, we incorporate an unsupervised local smoothness prior (TV loss) on the rendered depth at virtual views. 

\begin{figure}[t]
    \centering
    \includegraphics[width=\textwidth]{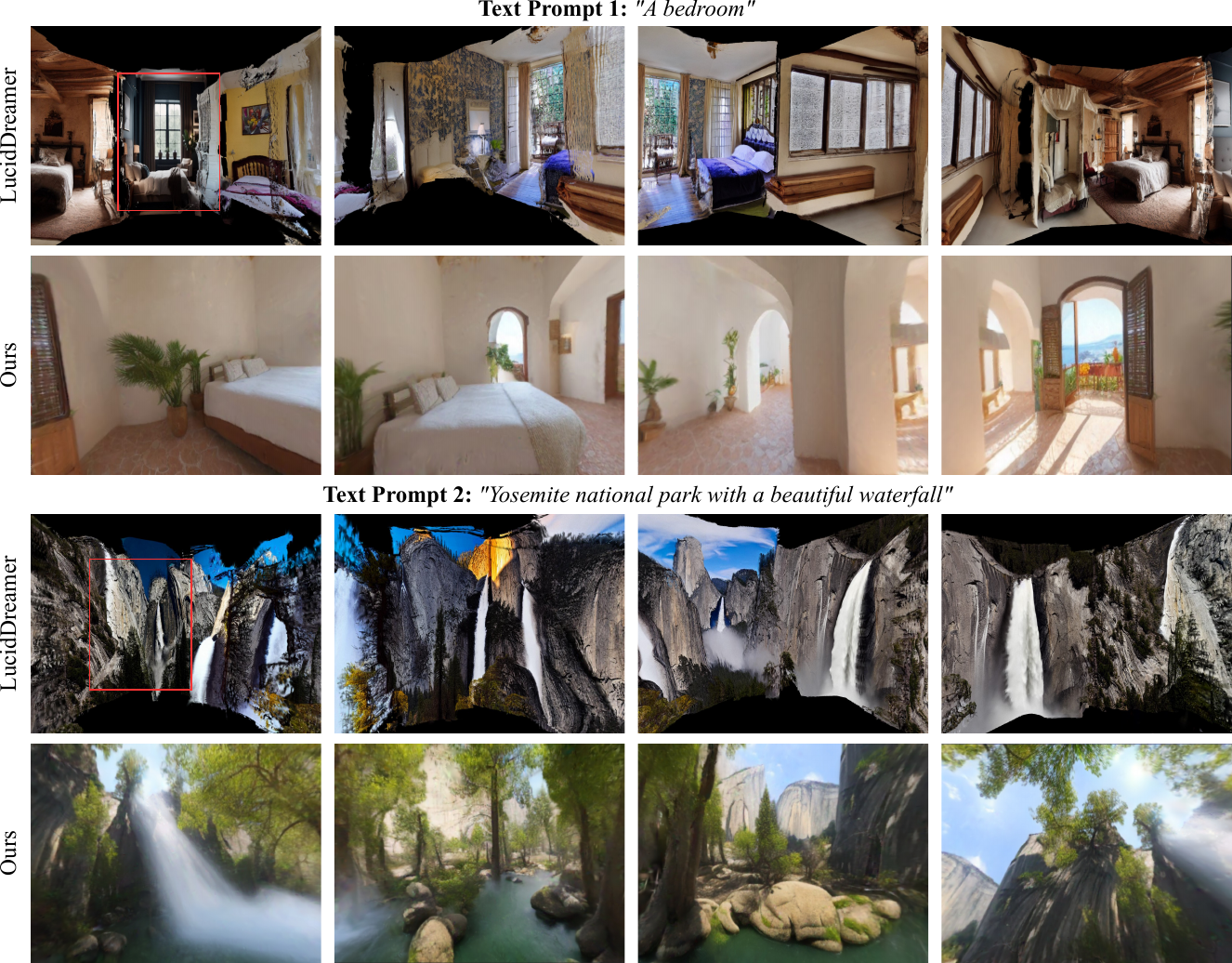}    
\caption{\textbf{Visual Comparisons.} We showcase 360$^\circ$ 3D scene generation. In each row, from left to right, displays novel views as the camera undergoes clockwise rotation in yaw, accompanied by slight random rotations in pitch and random translations. LucidDreamer~\cite{chung2023luciddreamer} hallucinates novel views from a conditioned image (indicated by a red bounding box) but lacks global semantic, stylized, and geometric consistency. In contrast, our method provides complete 360$^\circ$ coverage without any blind spots (black areas in baseline results), and shows globally consistent semantics.}
\label{fig:compare}
\end{figure}

\begin{table}[t]
  \caption{\textbf{Quantitative comparisons} between LucidDreamer and ours.}
  \label{tab:comparison}
  \centering
  \begin{tabular}{c@{\hspace{10pt}}c@{\hspace{10pt}}c@{\hspace{10pt}}c@{\hspace{10pt}}c@{\hspace{10pt}}c}
    \toprule
     & CLIP Distance$\downarrow$& Q-Align$\uparrow$ & NIQE$\downarrow$& BRISQUE$\downarrow$ &Runtime\\
    \midrule
    LucidDreamer~\cite{chung2023luciddreamer}  & 0.8900 & 3.0566 & 6.2305 & 51.9764 & 6min.15sec.\\
    Ours & \textbf{0.8732} & \textbf{3.1094} & \textbf{4.9165} & \textbf{38.3911} & 7min.20sec.\\
  \bottomrule
  \end{tabular}
\end{table}

\subsection{Optimization}
\label{sec:train_infer}
The entire pipeline for transforming a panorama into 3D can be supervised end-to-end through a composite of all the loss functions.
\begin{align}
    \mathcal{L} = \mathcal{L}_\textit{RGB} + \lambda_1 \cdot \mathcal{L}_\textit{sem} + \lambda_2 \cdot \mathcal{L}_\textit{geo}
\end{align}
Here, $\mathcal{L}_\textit{RGB}$ represents the photometric loss on projected perspective images, which consists of $\mathcal{L}_{1}$ and a D-SSIM term as described in~\cite{kerbl20233d}. $\mathcal{L}_\textit{sem}$ denotes semantic regularization, and $\mathcal{L}_\textit{geo}$ signifies geometric regularization. The weights $\lambda_1$ and $\lambda_2$ are set to 0.05 each, respectively.


\section{Experiments}


\subsection{Experiment Setting}

\paragraph{\textbf{Implementation.}}
Given the panorama generated from text, we resize it to $1024\times 2048$ in order to produce a dense enough point cloud by projecting per-pixel panoramic depth values into 3D space along the per-pixel camera ray directions, where the directions can be obtained according to the coordinate transformation in spherical panoramic imaging~\cite{rey2022360monodepth}. Since the panoramic depth map is predicted from the optimized geometric field, the values in local patches are guaranteed to be consistent and smooth, thereby resulting in point cloud in 3D that accurately captures the smooth geometry at the objects' surfaces. 
This setup enables effective 3D Gaussians initialization within a controllable, bounded area, eliminating redundant points in empty spaces. 
To circumvent the common floater issue in 3DGS rendering, we disable the densification process, enhancing the overall quality and consistency of the rendered scenes. 


\paragraph{\textbf{Baseline Methods.}}
Our work tackles the challenging problem of unconstrained 360$^\circ$ generation, including both indoor and outdoor scenarios. However, the works utilizing a bounded NeRF representation using score distillation do not work very well in this case.
Thus, the comparisons are conducted between DreamScene360 (ours) and the state-of-the-art LucidDreamer~\cite{chung2023luciddreamer}. We use the open-source codebase of LucidDreamer, which starts from a single image and a text prompt. The framework constructs a global point cloud by progressive inpainting into 360$^\circ$ views and then distills a set of 3D Gaussians. In our experiments, we set the input image for the LucidDreamer to be generated from Stable Diffusion~\cite{rombach2022high} v1.5 using the input text for a fair comparison.

\paragraph{\textbf{Metrics.}}
Since there is no ground truth in the generated 3D scenes, we utilize CLIP~\cite{radford2021learning} embedding distance, following previous works~\cite{xu2022neurallift,tang2023dreamgaussian}, to measure the text-image alignment as a mechanism to quantify the novel view rendering quality. Specifically, for each method, we render images with camera rotations and translations to mimic the immersive trajectory inside the 3D scenes. 
Additionally, we utilize multiple non-reference image quality assessment metrics. NIQE~\cite{mittal2012making} and BRISQUE~\cite{mittal2012no} are widely adopted non-reference quality assessment methods in measuring in-the-wild image quality. QAlign~\cite{wu2023q} is the state-of-the-art method in quality assessment benchmarks, which adopts a large multi-modal model fine-tuned on available image quality assessment datasets. We utilize the ``quality'' mode of QAlign, which focuses on the perceptual quality of image contents.

\subsection{Main Results}
\paragraph{\textbf{360$^\circ$ Scene Generation.}}
As shown in Fig.~\ref{fig:diverse}, our method can generate diverse 3D scenes in different styles with distinct contents, while preserving high-fidelity novel-view rendering ability and realistic scene geometry. These results showcase our method's generalization ability to diverse use cases, providing a user-friendly experience in realizing users' imaginations.

\paragraph{\textbf{Comparisons with Baseline Methods.}}

We show visual comparisons against LucidDreamer~\cite{chung2023luciddreamer} in Fig.~\ref{fig:compare}. LucidDreamer involves progressive inpainting to find agreement between multiple synthesized images and tries to fuse a point cloud that can be later distilled into 3D Gaussians. 
Their pipeline, which inpaints each patch separately based on the same text prompt, tends to produce repetitive results especially when generating complex scenes.
In comparison, our method delivers consistent results thanks to the intermediate panorama as a global 2D representation. We provide quantitative comparisons in Tab.~\ref{tab:comparison}, where the camera undergoes clockwise rotation
in yaw, accompanied by slight random rotations in pitch and random translations to capture 4 novel views roughly representing front, back, left, and right, along with two views of up and down by 90$^\circ$ rotation in pitch, to mimic the exploration in an immersive 360$^\circ$ panoramic 3D scene. 

\begin{figure}[t]
  \centering
  \includegraphics[width=0.99\linewidth]{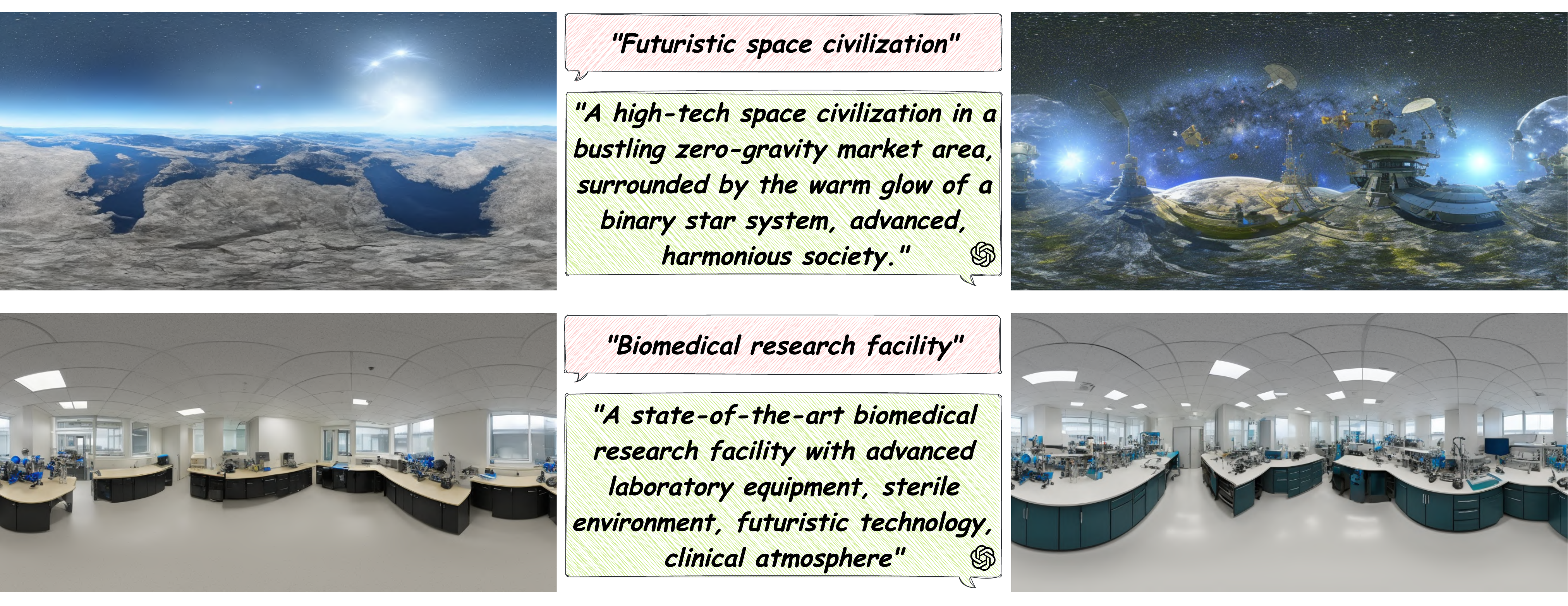}
  \caption{\textbf{Ablation of Self-Refinement.} We demonstrate that the self-refinement process greatly enhances the image quality by improving the text prompt. As shown in each row, the image on the left is generated using a simple user prompt, while a prompt augmented by GPT-4V obtains the image on the right. We observe that after the multi-round self-refinement, GPT-4V selects the one panorama with better visual quality, which provides solid support for the immersive 3D scene we further generate.}
  \label{fig:refine}
\end{figure}

For the \textit{bedroom} text prompt, LucidDreamer begins with a view featuring a modern style bedroom with one bed but hallucinates the novel views into multiple bedrooms, while the style of the imagined bedrooms diverges from that of the starting view. In the case of the \textit{Yosemite} text prompt, LucidDreamer merely replicates the waterfall seen in the initial view throughout. In conclusion, our results demonstrate global semantic, stylized, and geometric consistency, offering complete 360$^\circ$ coverage without any blind spots.

\subsection{Ablation Study}

\begin{figure}[t]
  \centering
  \includegraphics[width=0.99\linewidth]{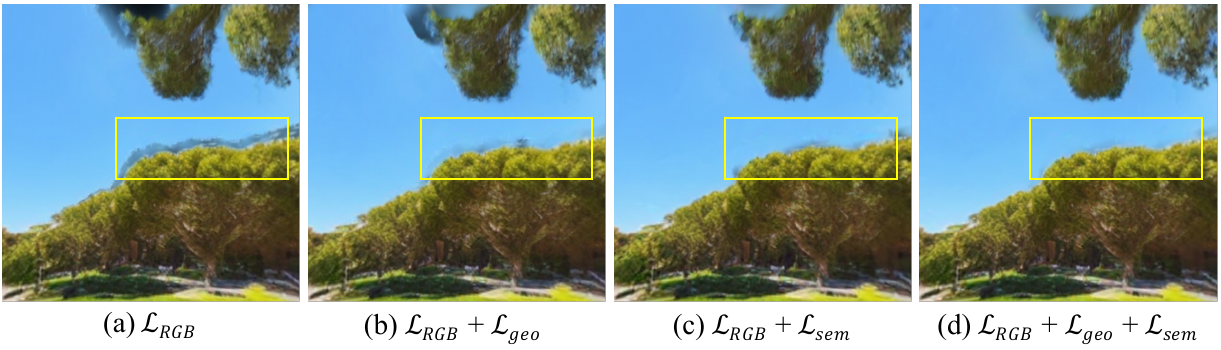}
  \caption{\textbf{Ablation of Optimization Loss.} We demonstrate the impact of Semantic and Geometric losses on the synthesized virtual cameras. (a) Utilizing photometric loss on camera views from a rendered panorama induces artifacts when altering rendered camera positions. (b) Implementing Geometric and (c) Semantic regularizations effectively reduces the artifacts originating from invisible views. (d) Integrating both regularizations yields the most optimal outcomes.}
  \label{fig:loss}
\end{figure}

\paragraph{\textbf{Self-refinement Process}}
We further evaluate the importance of the self-refinement process. In this part, we mainly focus on the generated panorama since panorama provides a holistic view of the 3D scene and provides an upper bound for the visual quality. As can be seen in Fig.~\ref{fig:refine}, we observe that using a simple text prompt usually delivers minimalist results with fewer details. With the help of GPT-4V, our self-refinement process enjoys prompt revision and quality assessment. This ability finally facilitates the selection of a more realistic and detailed panorama image among the draft image candidates. These functionalities are otherwise hard to achieve in previous baselines that do not have global 2D representations, and as a result, our results provide a much better visual appearance than baselines as shown in Fig.~\ref{fig:compare} and Tab.~\ref{tab:comparison}.

\paragraph{\textbf{Loss Function}} We investigate the importance of multiple loss functions we adopted in Fig.~\ref{fig:loss}. As shown in the figure, especially in the highlighted region, our full configuration of loss functions delivers the best visual quality when rendered from a novel viewpoint. In comparison, removing the geometric regularizations or semantic constraints will result in unpleasant artifacts.

\begin{figure}[t]
  \centering
\includegraphics[width=0.99\linewidth]{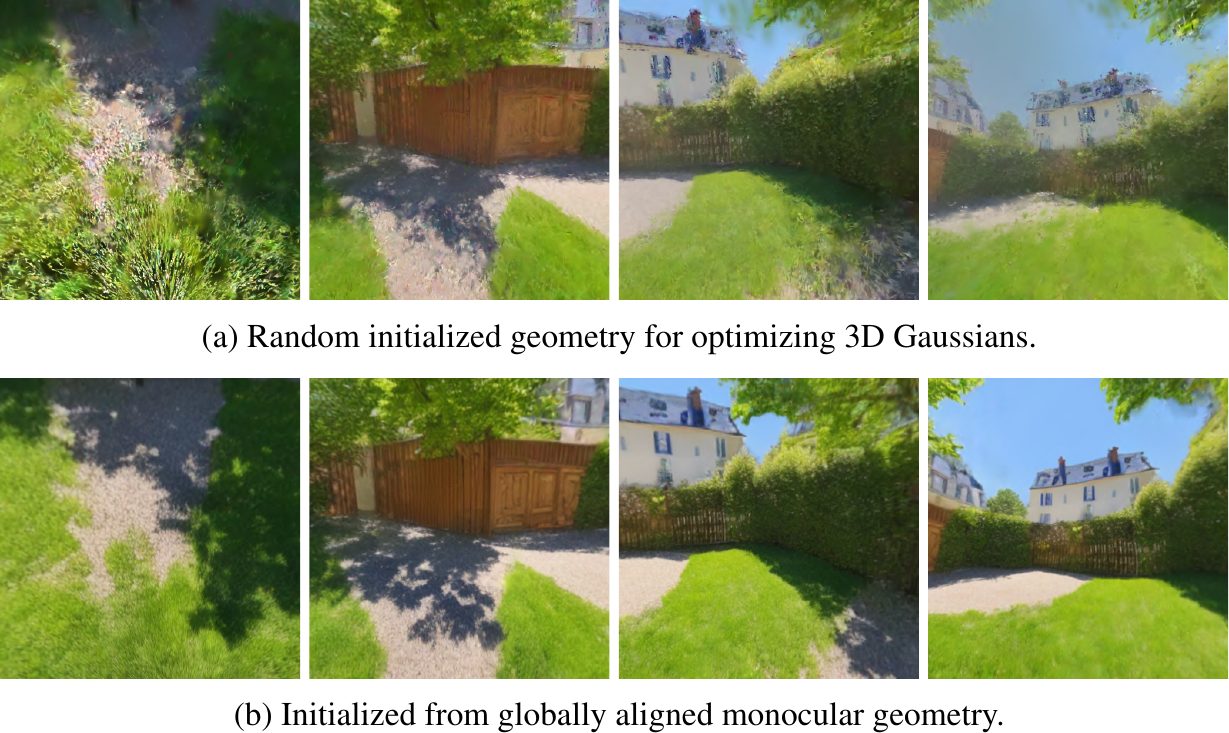}
\caption{\textbf{Ablation Study on 3D Initialization.} We present a comparative visualization of various initialization methods for 3D Panoramic Gaussian Splatting. In the absence of geometric priors of the scene (a), the optimized 3D Gaussian rendering yields plausible results in panoramic camera views. However, artifacts become evident when the rendered camera position is altered. To address this, we employ a monocular depth approach combined with learnable alignment factors and a geometric field (b), which ensures consistent alignment across multiple views.
  }
  \label{fig:init}
\end{figure}

\paragraph{\textbf{Initialization}} We showcase the importance of proper initialization in Fig.~\ref{fig:init}. As can be seen in the image, adopting our curated point initialization largely benefits the rendering quality of our 3D Panoramic Gaussian Splatting. An alternate solution involves adopting random initialization of the point cloud locations. This option, however, generates blurry results, mainly due to the challenge of learning proper point locations on the fly without ground truth geometry as supervision. Note that since our full model incorporates this high-quality point initialization, at the training stage, we can safely disable the adaptive density control of 3D Gaussians and, therefore, speed up convergence. Using random initialization of 3D Gaussians, on the other hand, has to rely on adaptive density control introduced in 3D Gaussian Splatting paper~\cite{kerbl20233d} (\textit{e.g.} densify, clone, prune) to move the 3D Gaussians to desired locations via backpropagation.


\section{Conclusion}
In this work, we present DreamScene360, a novel framework that unrestrictedly generates high-quality, immersive 3D scenes with complete 360$^{\circ}$ coverage in various styles from text inputs, regardless of their specificity level. Our pipeline leverages panorama as a middle ground, which provides us with a self-refinement opportunity by asking GPT-4V for quality assessment and prompt revision. Subsequently, we construct a geometric field that initializes 3D Gaussians. With the help of unsupervised geometric and semantic loss on virtual views, we lift the 2D panorama into panoramic 3D Gaussians. Experiments demonstrate our superiority against baseline methods in terms of global consistency and visual quality. However, our generation results are still limited by the default resolution ($512 \times 1024$) of the pretrained text-to-panorama diffusion model. Moving forward, we will explore generating 3D scenes at higher resolutions and even 4D dynamic scenes~\cite{li20244k4dgen} for a better, seamless, and immersive user experience.
\newpage
\section*{Acknowledgements}
We thank the Visual Machines Group (VMG) at UCLA, Visual Informatics Group at UT Austin (VITA), and ARL Intelligent Perception Branch for feedback and support. This project was supported by LUCI program under the Basic Research Office and partially supported by ARL grants W911NF20-2-0158 and W911NF-21-2-0104 under the cooperative A2I2 program. Z.W. is supported by the U.S. DEVCOM Army Research Laboratory and LUCI program, and an Army Young Investigator Award. A.K. is supported by a DARPA Young Faculty Award, NSF CAREER Award IIS-2046737, and Army Young Investigator Award.


%
%
\bibliographystyle{splncs04}
\bibliography{main}
\end{document}


\pagestyle{headings}
\mainmatter

\title{DreamScene360: Unconstrained Text-to-3D Scene Generation with Panoramic Gaussian Splatting} 


\titlerunning{DreamScene360}


\author{Shijie Zhou\inst{1}$^\star$\orcidlink{0000-0002-9018-7539} \and
Zhiwen Fan\inst{2}\thanks{Equal contribution.}\orcidlink{0000-0002-8302-7465
} \and
Dejia Xu\inst{2}$^\star$\orcidlink{0000-0001-8474-3095} \and
Haoran Chang\inst{1}\orcidlink{0009-0005-6333-239X} \and \\
Pradyumna Chari\inst{1}\orcidlink{0000-0002-9610-0350} \and
Tejas Bharadwaj\inst{1}\orcidlink{0009-0007-2370-8510} \and
Suya You\inst{3}\orcidlink{0000-0002-6387-7024} \and \\
Zhangyang Wang\inst{2}\orcidlink{0000-0002-2050-5693} \and
Achuta Kadambi\inst{1}\orcidlink{0000-0002-2444-2503}
}

\authorrunning{S.~Zhou et al.}


\institute{University of California, Los Angeles \and
University of Texas at Austin \and
DEVCOM Army Research Laboratory 
}
\maketitle

\section*{Supplementary Content}
This supplement is organized as follows:

\begin{itemize} 
    \item Section~\ref{sec:algorithm} contains the algorithmic details of our proposed method.
    \item Section~\ref{sec:gpt-4v} contains the experimental details of our self-refined panorama generation module with GPT-4V.
    \item Section~\ref{sec:environment} contains the details of the environment setup for our experiments.   
    \item Section~\ref{sec:results} contains additional qualitative results.
    \item Section~\ref{sec:baseline} contains additional results on baseline comparison. 
    \item Section~\ref{sec:additional_ablation} contains additional ablation studies.

\end{itemize}

\section{Algorithmic Details} \label{sec:algorithm}

Given the input text prompt with any level of specificity, we utilize GPT-4V to perform prompt revision and image quality assessment inspired by Idea2Img [\cnohref{69}]. Different from their original setup which only works for generating ordinary 2D images, our multi-round self-refinement module is incorporated with the pretrained text-to-360$^\circ$ panoramic image diffusion model [\cnohref{62}] which guarantees the generated panorama images fully satisfy the equirectangular format for an ominidirectional image representation. An upsampling process by bilinear interpolation is followed by obtaining the best candidate panorama image result, to ensure a sufficiently dense point cloud for the 3D Gaussians initialization. In practice, we resize the image from 512 $\times$ 1024 to 1024 $\times$ 2048. For the geometric field optimization, we use one set of Icosphere projection: 20 perspective images with FoV 80$^\circ$ and resolution 512 $\times$ 512 to cover the whole panoramic sphere, following [\cnohref{50}] and [\cnohref{61}]. For the optimization of panoramic Gaussian splatting, the training set consists of 12 sets of perspective images of the Icosphere projection with random rotation in radian. Additionally, different from 3D Gaussian Splatting (3DGS) [\cnohref{25}], we turn off the Gaussians densification as our dense point cloud initialization has already accurately captured the surfaces in the scene and redundant 3D Gaussians produced from densification will result in floaters near the surfaces and in the empty space. Algorithm~\ref{alg:dreamscene360} illustrates the pseudo code of our DreamScene360.

\begin{algorithm}[h]
	\caption{DreamScene360}
	\hspace*{\algorithmicindent} \textbf{Input:} ${TextPrompt_{0}}$
	\begin{algorithmic}[1]
        \State  $i \gets 0$ \Comment{Iteration Counter}
        \While{not maximum iteration}
        \State ${TextPrompt_{i}} \gets \text{GPT-4V}({TextPrompt_{i}})$ \Comment{Prompt Revision}
        \State ${P_{i}} \gets \text{PreTrainedDiffusionModel}({TextPrompt_{i}})$ \Comment{Panorama Image Candidate}
        \State ${S_{i}} \gets \text{GPT-4V}(P_{i})$ \Comment{Candidate Score}
        \State ${TextPrompt_{i+1}} \gets \text{GPT-4V}({TextPrompt_{i}})$ \Comment{Self-Refinement}
        \EndWhile
        \State ${P} \gets MaxScore(P_{s})$ \Comment{Select Panoroma with Best Score}
        \State ${P} \gets UpSample(P)$ \Comment{Upsample by bilinear interpolation}
        \State ${I_{N}}, {C_{N}} \gets \text{Projection}({P})$ \Comment{Get Perspective Images and Camera Poses}
        \State ${D_{N}^{Mono}} \gets \text{DepthEstimator}({I_{N}})$ \Comment{Get Monocular Depth}
        \State $\text{GeometricField} \gets {D_{N}^{Mono}} + v$ \Comment{Learn GeometricField with view direction}
        \State ${PointCloud} \gets \text{GeometricField}$ \Comment{Global Depth Alignment}
        \State ${Gaussians} \gets {PointCloud}$ \Comment{3D Gaussians Initialization}

		\While{not converged}
		\State ${I}, {C}, {D} \gets \text{GetTrainingView()}$ \Comment{Given Image, Camera Pose, Depth Map}
		\State  $\hat{{I}}, \hat{{D}} \gets \text{Rasterizer}({Gaussians}, {C})$  \Comment{Rendered Image, Remdered Depth Map}
		\State ${L} \gets \text{Loss}({I}, \hat{{I}}) + \lambda_{d}\text{Loss}({D}, \hat{{D}})$ \Comment{Loss Calculation}
        \If{PerturbationIterationStages}
            \State ${I}, {C'} \gets \text{GetVirtualView()}$ \Comment{Virtual Camera Pose, Imagep}
            \State  $\hat{{I'}}, \hat{{D'}} \gets \text{Rasterizer}({Gaussians}, {C'})$ 
            \State \Comment{Rendered Perturbation Image, Rendered Perturbation Depth Map}
            \State  ${D'} \gets \text{DepthEstimator}(\hat{{I'}})$ \Comment{Perturbation Depth Map}
            \State  ${[CLS]} \gets \text{FeatureExtractor}({I})$ \Comment{Reference DINO Token}
            \State  ${[CLS]'} \gets \text{FeatureExtractor}({I'})$ \Comment{Perturbation DINO Token}
            \State ${L} \gets {L} + \lambda_{f}\text{Loss}({[CLS]}, {[CLS]'}) + \lambda_{d}\text{Loss}({D'}, \hat{{D'}})$ \Comment{Loss Calculation}
        \EndIf
        \State ${Gaussians} \gets \text{Adam}({L})$ \Comment{Optimization}
        \EndWhile
	\end{algorithmic}
	\label{alg:dreamscene360}
\end{algorithm}


\section{Self-Refinement for Panoramic Image Generation}
\label{sec:gpt-4v}

We demonstrate the workflow of our self-refinement module using GPT-4V for multi-round text prompt revision and image quality evaluation for our text to 360$^\circ$ panoramas generation. In our implementation, we fix the trigger word \textit{360-degree panoramic image} as a compulsory part of the input text prompt for the Text2Pano diffusion model [\cnohref{62}]. We also adopt negative prompts to ensure the generation quality: \textit{lowres, bad anatomy, bad hands, text, error, missing fingers, extra digit, fewer digits, cropped, worst quality, low quality, normal quality, jpeg artifacts, signature, watermark, username, blurry}.

Take the input text prompt \textbf{"Castle on the hill"} as an example, we showcase one round of refinement and assessment by GPT-4V as follows:

\begin{mdframed}
    iteration 0

\noindent prompt: 360-degree panoramic image, Majestic castle perched atop a lush green hill, medieval architecture, serene sky backdrop. 

\vspace{5pt}

\noindent iteration 1

\noindent prompt: 360-degree panoramic image, Ancient stone castle standing on a windswept hill, ominous clouds gathering above, twilight setting. 

\vspace{5pt}

\noindent iteration 2

\noindent prompt: 360-degree panoramic image, Grand castle with towering spires on a sunlit hill, surrounded by wildflowers, clear blue sky.

\vspace{5pt}

\noindent iteration 3

\noindent prompt: 360-degree panoramic image, Imposing castle on a steep hill, silhouetted against a fiery sunset, foreboding atmosphere.

\vspace{5pt}

\noindent iteration 4

\noindent prompt: 360-degree panoramic image, Enchanted castle nestled on a misty hill, fairy tale inspired, surrounded by an enchanted forest, dawn light. 

\noindent ===============

\noindent End of round 0:

\noindent user prompt: Castle on the hill.

\noindent select history: Round selection: 2.

\noindent To evaluate how well these images match the user's idea of a "Castle on the hill," I will consider the presence of a castle, its placement on a hill, and the overall composition that aligns with this concept.

Image 0: There is a castle on a hill in the distance, surrounded by a green landscape and a clear sky. This image closely matches the idea. Score: 8/10

Image 1: This image shows a hill, but there is no castle present. It does not match the idea well. Score: 2/10

Image 2: There is a castle with multiple spires on a hill with a sunny sky and some flowers in the foreground. This image aligns well with the idea. Score: 9/10

Image 3: A castle is present on a hill with a dramatic sunset sky. This image also aligns well with the idea. Score: 8/10

Image 4: This image shows a hill with trees and fog but lacks a castle. It does not match the idea. Score: 1/10

\noindent ===============

\noindent Considering the scores and the adherence to the idea of a "Castle on the hill," the image with the best overall quality is: 2

\end{mdframed}

\begin{figure}[h]
    \centering
    \includegraphics[width=0.65\textwidth]{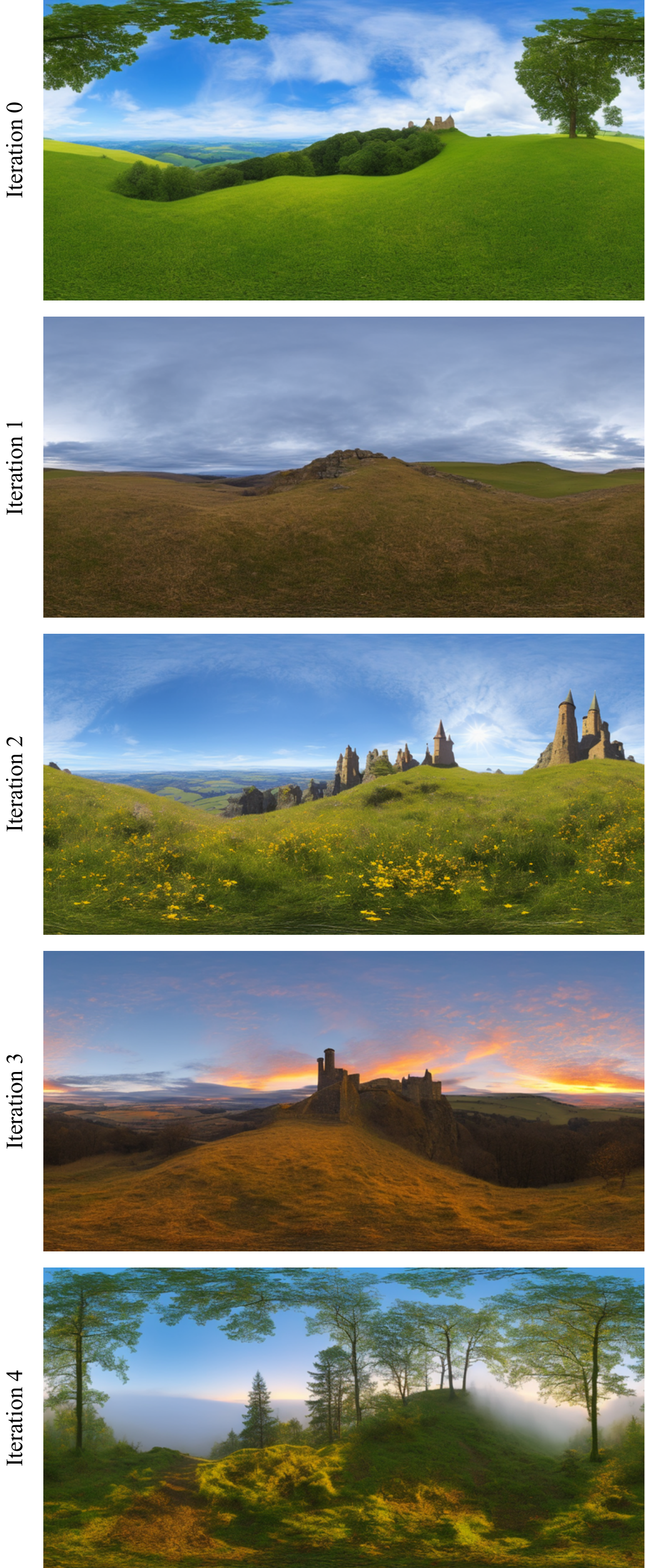}
    \caption{\textbf{Self-refinement by GPT-4V.} Input text prompt: "Castle on the hill".}
    \label{fig:prompt_refinement}
\end{figure}

Correspondingly, Fig.~\ref{fig:prompt_refinement} presents the panoramic images generated at each iteration. It is evident that the selection made by GPT-4V aligns with both input text description and human evaluation outcomes.

\clearpage

\section{Environment Setup:}
\label{sec:environment}
We describe the detailed setup that has been used in our experiments and runtime tests:
\begin{itemize}
    \item CUDA version: 11.8
    \item PyTorch version: 2.0.1
    \item GPU: single NVIDIA GeForce RTX 4090
    \item CPU: 13th Gen Intel(R) Core(TM) i9-13900KF   3.00 GHz
\end{itemize}

Generally, one text-to-3D generation experiment takes no more than 16GB GPU memory.

\section{Additional Qualitative Results} 
\label{sec:results}
\begin{figure}[h]
    \centering
    \includegraphics[width=0.9\textwidth]{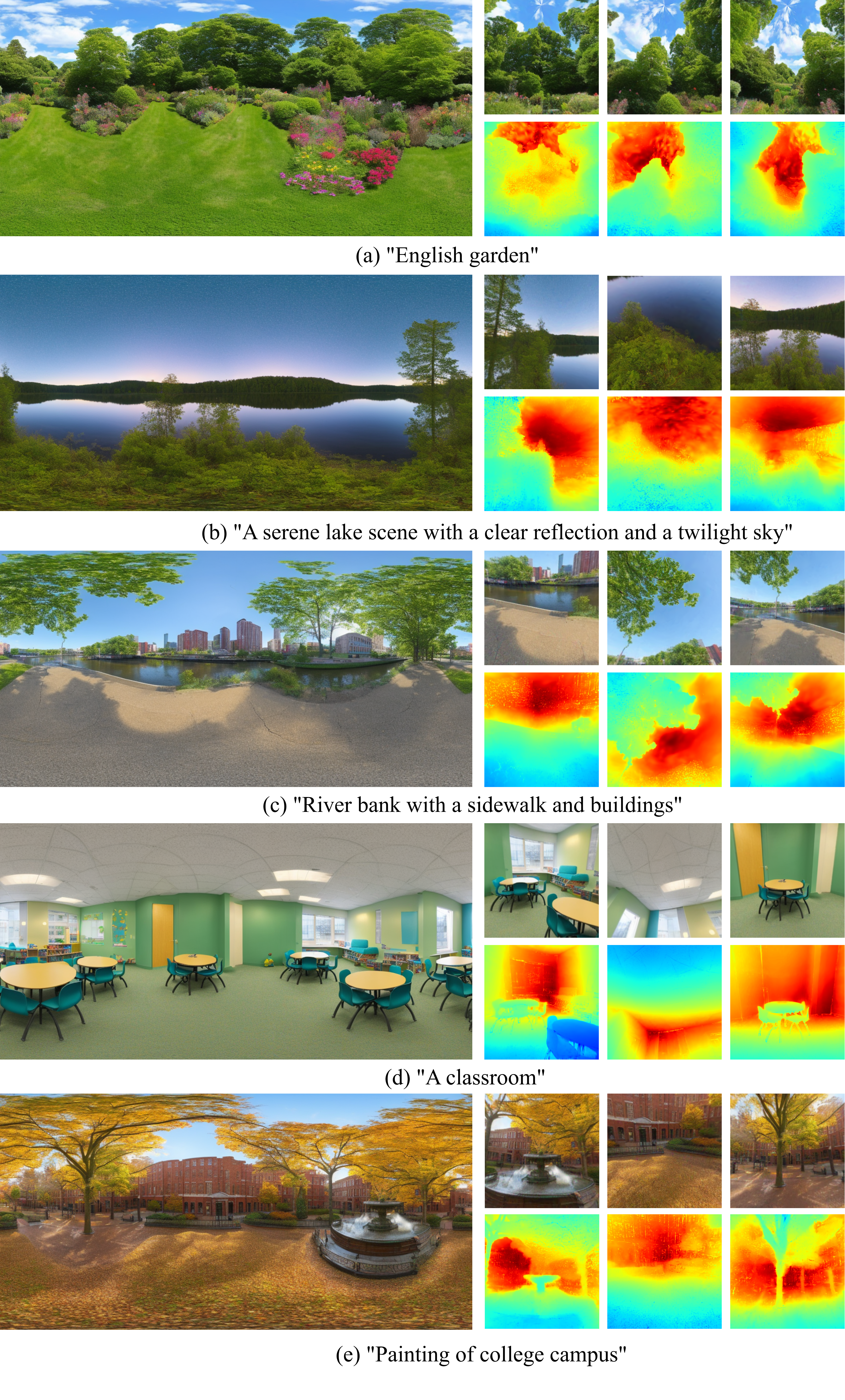}   
\caption{\textbf{Additional Qualitative Results on Diverse Generation.}}
\label{fig:diverse}
\end{figure}

We present further qualitative findings in Fig.~\ref{fig:diverse}, showcasing the generated 2D panorama, along with the rendered perspective images and depth maps from our panoramic Gaussian radiance field. The results demonstrate that our DreamScene360 is adept at generating varied 3D scenes with precise geometry, including both unbounded outdoor scenes and bounded indoor scenes with different kinds of styles. Additionally, we highly recommend viewers explore our video demonstrations for an immersive experience of "flying" (freely translating and rotating) within the generated 3D scene.

\section{Additional Results on Baseline Comparison} \label{sec:baseline}
\subsection{Text-to-3D Generation}
\begin{figure}[h]
    \centering
    \includegraphics[width=\textwidth]{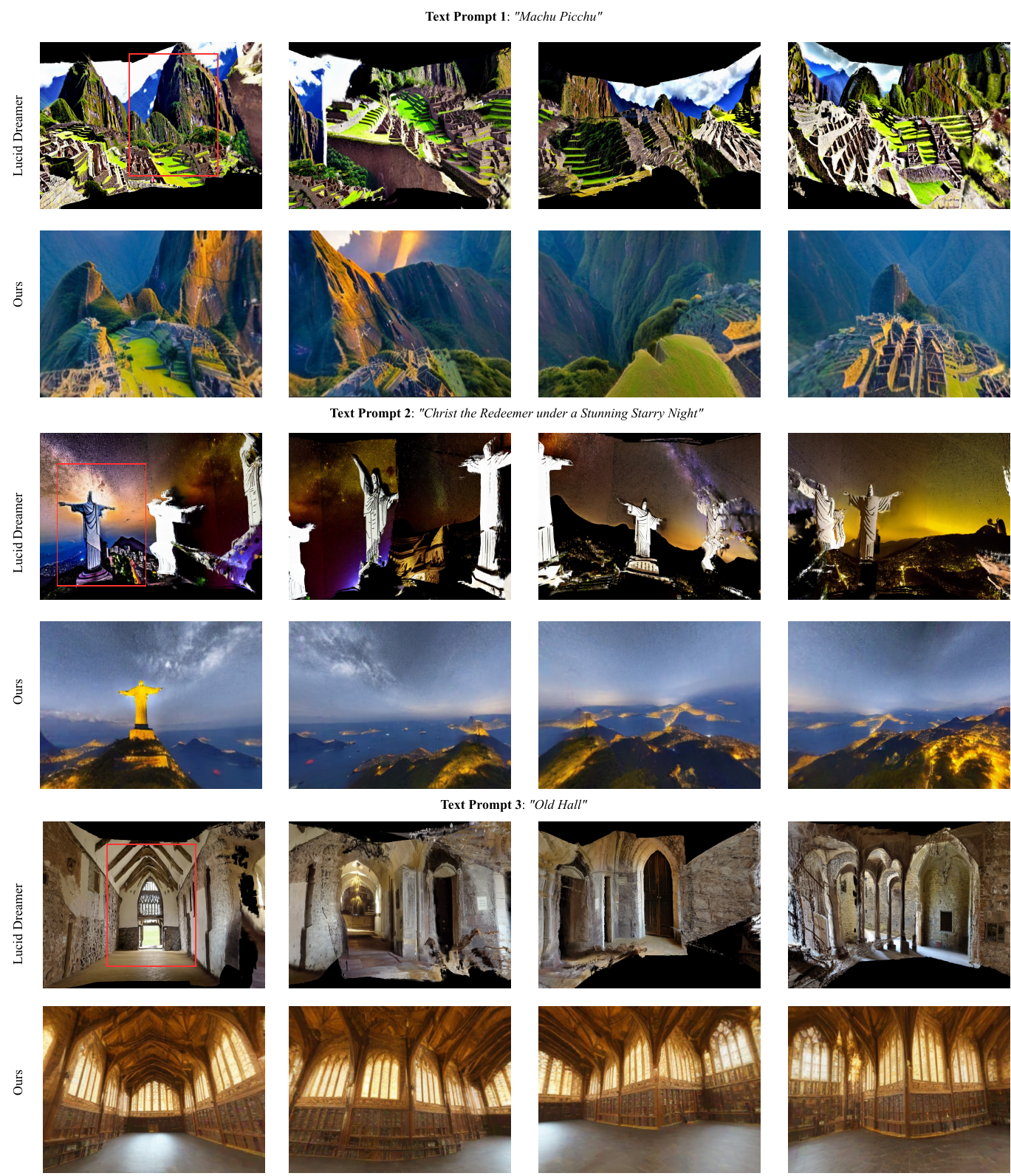}    
\caption{\textbf{Additional Baseline Comparisons.} DreamScene360 v.s. LucidDreamer [\cnohref{7}] regarding text-to-3D scene generation.}
\label{fig:compare_sup}
\end{figure}

We present more visual results compared to LucidDreamer [\cnohref{7}] in Fig.~\ref{fig:compare_sup}, captured with a clockwise rotation in yaw and random translation. Unlike LucidDreamer, which exhibits repetitive patterns in its first (\textit{Machu Picchu}) and second (\textit{Christ the Redeemer}) examples, our method maintains consistent photographic quality throughout the scene. Furthermore, while LucidDreamer's prior front views diverge from its generated the back views, our approach ensures a more coherent visual transition. Notably, in the third example (\textit{Old Hall}), LucidDreamer's outputs demonstrate inconsistencies in geometry and feature black spots under some random translations, showcasing our advantages in maintaining geometric consistency.

\subsection{Panorama-to-3D Lifting}
\begin{figure}[h]
\begin{center}
\includegraphics[width=1.0\linewidth]{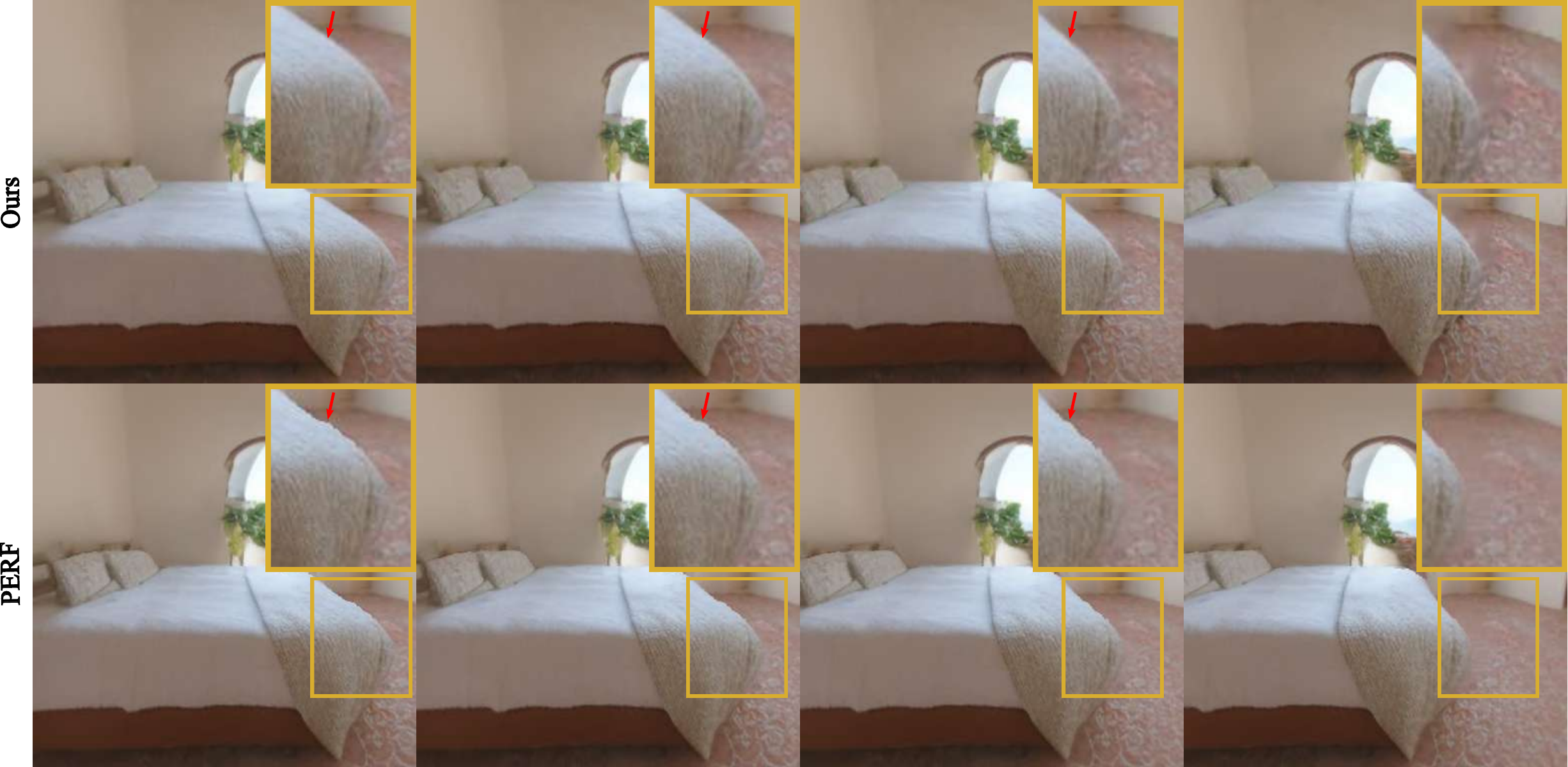}
\caption{\textbf{Additional Baseline Comparisons.} DreamScene360 v.s. PERF [\cnohref{61}] regarding panorama-to-3D lifting. We show comparable visual results on novel views to PERF. In addtion to that, our work shows better geometry on the edges.}
\label{fig:occlusion}
\end{center}
\end{figure}

PERF [\cnohref{61}] is a great NeRF-based panorama-to-3D lifting work demonstrated on indoor scenes, whereas DreamScene360 is for text-to-3D scene generation using Gaussian Splatting.
Thanks to our geometric initialization and refinement step, we can lift panoramic images showcasing a wide range of scene contexts and artistic styles (e.g., indoor, outdoor, cartoon) to 3D. The comparison with PERF is not included in the main paper since PERF is not a 3DGS-based text-to-3D scene generator. However, since the panorama-to-3D part is comparable, we present comparison results using the same panorama image as input. Fig.~\ref{fig:occlusion} shows ours achieving more accurate geometry (PERF has zigzaging artifacts at bed edge, see red arrows) because PERF directly performs inpainting on 2D images to train a NeRF, while we use rendered images and both semantic and geometric supervision to guide optimization in 3D. 
Tab.~\ref{tab:baseline_comparison_perf} shows that ours is superior across all metrics of rendering quality, while being significantly faster.

Besides rendering quality, occlusion is a fundamental problem in single view panorama to 3D reconstruction. 
DreamScene360 starts from an omnidirectional panoramic image and lift it to 3D using global depth cues. 
The proposals in ``Distilling Semantic Similarities'' and  ``Regularizing Geometric Correspondences'' on unseen views are specifically designed to rectify occlusions.
We compare our method with PERF by translating right from given single training view (leftmost Fig.~\ref{fig:occlusion}), and we observe that our 3D approach achieves similar degree of occlusion compared to PERF's 2D inpainting (see Fig.~\ref{fig:occlusion}, rightmost column) but with 4.6$\times$ faster training (Tab.~\ref{tab:baseline_comparison_perf}).
 

\begin{table}[t]
  \caption{Quantitative results: ours vs. PERF [\cnohref{61}] }
  \label{tab:baseline_comparison_perf}
    \centering
    \resizebox{0.65\linewidth}{!}{
      \begin{tabular}{@{}lcccc@{}} 
        \toprule
         & PSNR$\uparrow$ & SSIM$\uparrow$ & LPIPS$\downarrow$ & Training Time\\
        \midrule
        PERF & 36.881 & 0.957 & 0.074 & 44min.01sec. \\
        Ours & \textbf{40.179} & \textbf{0.970}& \textbf{0.035} & 9min.32sec.  \\
        \bottomrule
      \end{tabular}
    }
\end{table}


\section{Additional Ablation Studies} \label{sec:additional_ablation}
\subsection{Ablations of Virtual Camera Perturbations.}
\begin{figure}[t]
    \centering
    \includegraphics[width=1.0\textwidth]{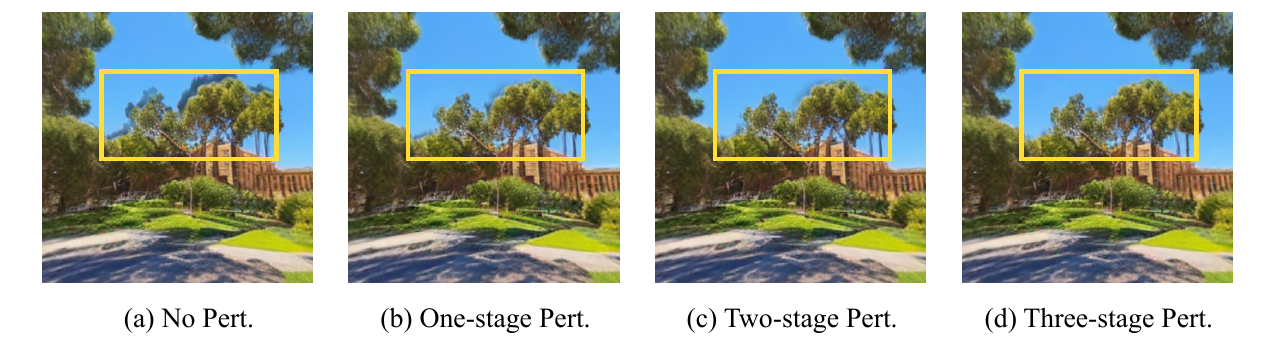}   
\caption{\textbf{Ablation of Virtual Camera Perturbations.} We show the influence of adding levels of virtual camera perturbations; the results improve progressively with more stages involved.}
\label{fig:ablation_perturbations}
\end{figure}

To emulate the camera's movement from the original view point, we introduce virtual cameras with 3-stage progressive perturbations with 3 levels. In our experiments, we start the 3-stage progressive camera perturbation from 5400 iterations during the training, the interval of each stage is adjustable, and we set it to 1200 iterations in practice. With the same observation in [\cnohref{25}], most of the scenes at 9000 iterations can already achieve very high quality. We study the influence of perturbations on the results. Fig.~\ref{fig:ablation_perturbations} shows the qualitative results with different level of perturbations while keeping the same total number of iterations.

\subsection{Ablations of FOV.}
\begin{figure}[h]
\begin{center}
\small
\includegraphics[width=1.0\linewidth]{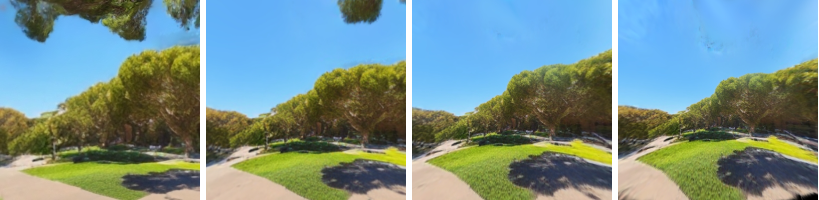}
\caption{\textbf{Ablation of Inference FOV.} FOV set to 36$^{\circ}$, 80$^{\circ}$, 102$^{\circ}$, 118$^{\circ}$ (left to right).}
\label{fig:fov}
\end{center}
\end{figure}

During inference, the field of view (FOV) is one of the adjustable parameters. In practice, we set a large camera FOV (80$^\circ$) when rendering. Since we use the pinhole camera model, a large FOV may result in distortion. We show results with different FOVs in Fig.~\ref{fig:fov}, where it is obvious that a smaller FOV alleviates these distortions.

\clearpage
%
%